\RequirePackage{fix-cm}
\documentclass[smallcondensed]{svjour3}     % onecolumn (ditto)
\smartqed  % flush right qed marks, e.g. at end of proof
\usepackage{graphicx}
\usepackage{amsmath}
\usepackage{amsfonts}
\usepackage{amssymb}
\usepackage{graphicx}
\usepackage{tikz,pgfplots}
\usetikzlibrary{shapes,snakes}
\usepackage{epstopdf}
\usepackage{color}%
\usepackage{subfigure}
\usepackage{enumerate}
\usepackage[sort&compress,numbers]{natbib}
\begin{document}

\title{A Novel Adaptive Deep Network for Building Footprint Segmentation%\thanks{Grants or other notes
\\
}

\titlerunning{A Novel Adaptive Deep Network for Building Footprint Segmentation}        % if too long for running head

\author{A. Ziaee         \and R. 
Dehbozorgi     \and % Professor (FH) PD Dr. 
       M. D\"{o}ller
        %etc.
}


\institute{Amir Ziaee \at The University of Applied Science FH Kufstein	
              \email{amir.ziaee@fh-kufstein.ac.at}           %  \\
           \and
           Raziyeh Dehbozorgi \at Iran University of Science and Technology
            \email{r.dehbozorgi2012@gmail.com} 
           \and
           Mario D\"{o}ller \at  The University of Applied Science FH Kufstein	
              \email{Mario.Doeller@fh-kufstein.ac.at} 
}

\date{Received: date / Accepted: date}

\maketitle

\begin{abstract}

Building footprint segmentations for high resolution images are increasingly demanded for many remote sensing applications. By the emerging deep learning approaches, segmentation networks have made significant advances in the semantic segmentation of objects. However, these advances and the increased access to satellite images require the generation of accurate object boundaries in satellite images. In the current paper, we propose a novel network based on Pix2Pix methodology to solve the problem of inaccurate boundaries obtained by converting satellite images into maps using segmentation networks in order to segment building footprints. To define the new network named G2G, our framework includes two generators where the first generator extracts localization features in order to merge them with the boundary features extracting from the second generator to segment all detailed building edges. Moreover, different strategies are implemented to enhance the quality of the proposed networks' results, implying that the proposed network outperforms state-of-the-art networks in segmentation accuracy with a large margin for all evaluation metrics. The implementation is available at https://github.com/A2Amir/A-Novel-Adaptive-Deep-Network-for-Building-Footprint-Segmentation.


\end{abstract}
\section{Introduction}
\begin{figure}[ht]
\begin{center}
\subfigure[Satellite Image]{
\resizebox*{3.7cm}{!}{\includegraphics{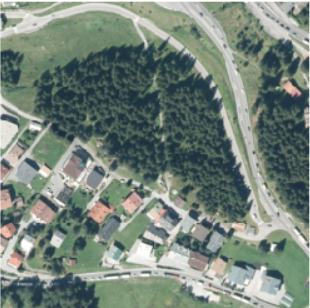}}}\hspace{3pt}
\subfigure[Ground Truth ]{
\resizebox*{3.7cm}{!}{\includegraphics{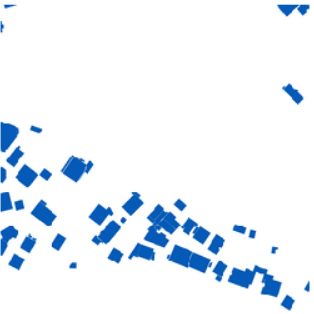}}}
\subfigure[Prediction]{
\resizebox*{3.7cm}{!}{\includegraphics{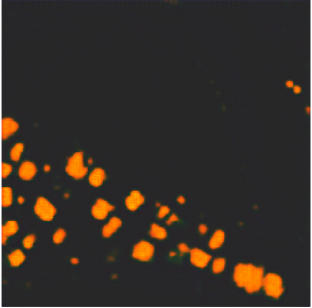}}}
\caption{Prediction with inaccurate boundaries.}
\label{fig1}
\end{center}
\end{figure}

Accurate segmentation is an important issue in the field of computer vision. To have an exact segmentation, every pixel of an image should be classified into multiple segments. Although, regardless of knowing objects, human visual system can segment all unknown objects of an image such as a satellite image. Computers get into a challenge to conduct such a task. For instance, extracting buildings automatically from an urban image is complex, since roof textures are different in terms of shape and size. Also, the contrast between buildings and its environments is very low \cite{29}. Therefore, efficient extraction of building footprints remains a challenge.
\\
Building detection and footprint extraction with all salient features are of high interest for many applications such as urban planning, building monitoring, sociology, and disaster emergency response. Regarding the variety of the building materials and scales, buildings in aerial/satellite imagery are depicted significantly different. In spite of the abundant research to detect and extract building footprints, a few algorithms have been investigated to deal with the problem of inaccurate edges \cite{11}. Therefore, the quality of results such as precision and resolution of boundaries remain as valid concerns.
\\
Semantic segmentation, representing as one of the granularity levels within the segmentation process, classifies each pixel into a specific class. This level of segmentation is applied in many sorts of analysis such as satellite imagery analysis, recognition of image copies, and human-computer interaction \cite{app1,app2,app3}. However, the challenge remains on the correct detection and classification of the individual objects for producing predictions with accurate boundaries \cite{app1,11}. \\
Fully Convolutional Networks (FCNs) was firstly introduced for semantic segmentation \cite{32}. Although, this has been applied to segment satellite images, the limitation of low-resolution predictions leads to further research to obtain better models. Therefore, a number of techniques have been proposed to address this limitation aiming at generating high-resolution and accurate boundaries. In this regard, the first U-Net was built for the segmentation of biomedical images \cite{34}, which was extended for many other applications in the field of segmentation, in order to overcome the aforementioned issues \cite{35,61}.  
\\
Although, deep learning networks play the most prominent role in the area of semantic segmentation, when analyzing satellite images, they have shown some drawbacks in producing boundaries (Fig. \ref{fig1}). 
To conquer the problem of inaccurate prediction of boundaries in satellite images, this study proposes a deeper network architecture using two generators based on Pix2Pix network \cite{35}, being developed based on the CGAN \cite{45}. Accordingly, current study covers the followings: \\
 \\
\begin{itemize}
\item[$\bullet$] Focusing on Pix2Pix network to perform detailed experiment design for segmentation of buildings. A proof of the concept, implying that using two generators are more powerful than one. 

\item[$\bullet$] Representing a new network with a new architecture to improve the accuracy of semantic segmentation networks, which enables us to demonstrate an increased accuracy of approximately 10 percent in the segmentation of building footprints in satellite imagery processing. 

\item[$\bullet$] Implementing four-evaluation metrics to prove that our approach significantly exceeds Pix2Pix network and the current state-of-the-art networks in term of accuracy at specified validation without any hyper-parameter optimization.
\end{itemize}
\section*{Related works}
Extracting the momentous features of building footprint are of great importance for creation of an appropriate map with a high local accuracy. Edges are among those features, being rarely investigated due to the various buildings' shapes and scales. Despite the efforts that have been put into developing methodologies based on deep neural networks, to provide an automatic extraction of the building footprints, they are not still producing satisfactory results. \\
Some efforts have been made to solve the problem of preserving semantic segmentation boundaries. Bittner et al. \cite{12} introduced an approach to automatically generate a full resolution binary building mask without any assumptions on the scale and shape of the buildings using a Digital Surface Model (DSM) and an FCN architecture. Their solution includes two main steps: i) Training FCN on a large set of patches consisting of normalized DSM (nDSM) as inputs and available ground truth building mask as target outputs. ii) Considering the generated predictions from FCN as unary terms for a fully connected Conditional Random Field (FCRF), which enables them to achieve a final binary building mask. This work was later improved by demonstrating an end-to-end fused-FCN in which the fusion of several networks including three-band (RGB), panchromatic (PAN), and nDSM results in high resolution images \cite{13}.
Additionally, an algorithm based on the combination of the robust Classification-convolution Neural Networks (CNN) with an Active Contour Model (ACM) have been introduced to improve the accuracy of current building edge extraction \cite{14}. ACM can be accounted as a useful tool to elevate the accuracy, in case, the footprints of a building are missed in the CNN classification.\\
To increase the accuracy of semantic segmentation of high-resolution images, even more,  Xiaoye Wang et al. \cite{15} also improved Pix2Pix network by adding a controller to have a new Pix2Pix model called ePix2Pix to progress the classification performance and create segmentations that more efficiently match with the ground truths in terms of shape. Another similar work was done by Schuegrafwe and Bittner \cite{16} to develop a deep learning-based algorithm for DSMs and spectral images (PAN and multi-spectral) fusion. They utilized an end-to-end U-Net to combine depth and spectral information within two parallel networks. This led to a combination of the features in the late phase to obtain binary building masks using a residual block of the neural network.
\\
Zhu et al. \cite{17} developed a new Multi Attending Path Neural Network (MAP-Net) for accurate extraction of building footprints and precise boundaries on multiple levels. They extracted boundary and semantic information through two different networks and merged this information in later stages. 
Bischke et al. \cite{11} achieved an advancement in this area by proposing an uncertainty weighted and cascaded multi-task loss based on distance transform accompanied, by a deeper network architecture to improve semantic segmentation predictions.\\ 
Based on the previous studies, this work design a pre-processing algorithm and a strategy  using two different networks and modify Pix2Pix architecture. The new resulting Pix2Pix network termed as G2G network allows us to achieve an improvement of the semantic segmentation in respect to accuracy.
\section{Methodology: G2G} 
\begin{figure}[ht]
\begin{center}
\subfigure[Ground truth]{
\resizebox*{3.7cm}{!}{\includegraphics{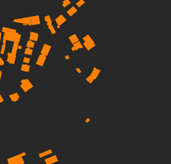}}}\hspace{3pt}
\subfigure[Boundaries ]{
\resizebox*{3.7cm}{!}{\includegraphics{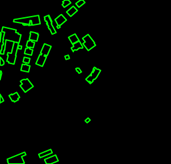}}}
\subfigure[Combination]{
\resizebox*{3.7cm}{!}{\includegraphics{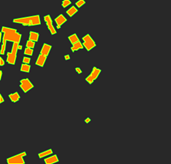}}}
\caption{Extracting and re-adding the contours to objects of the ground truths.}
\label{fig2}
\end{center}
\end{figure}

\begin{figure}[ht]
\begin{center}
\subfigure[The generator]{
\resizebox*{3.7cm}{!}{\includegraphics{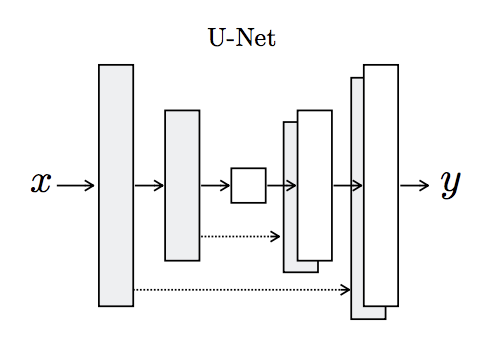}}}
\subfigure[The Patchgan discriminator]{
\resizebox*{7.7cm}{!}{\includegraphics{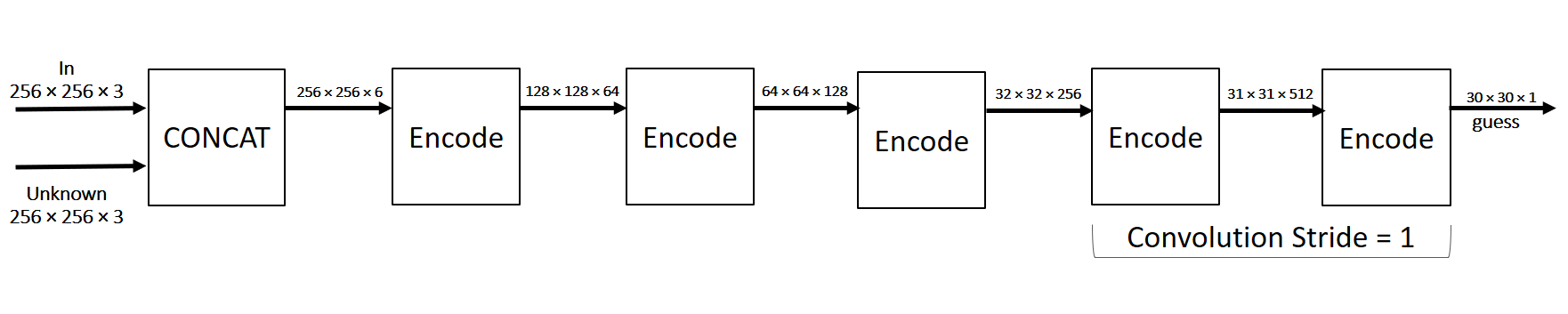}}}
\caption{Pix2Pix structures of the generator and Patchgan discriminator \cite{34}. }
\label{fig3}
\end{center}
\end{figure}

\begin{figure}[ht]
\begin{center}
\resizebox*{12.5cm}{!}{\includegraphics{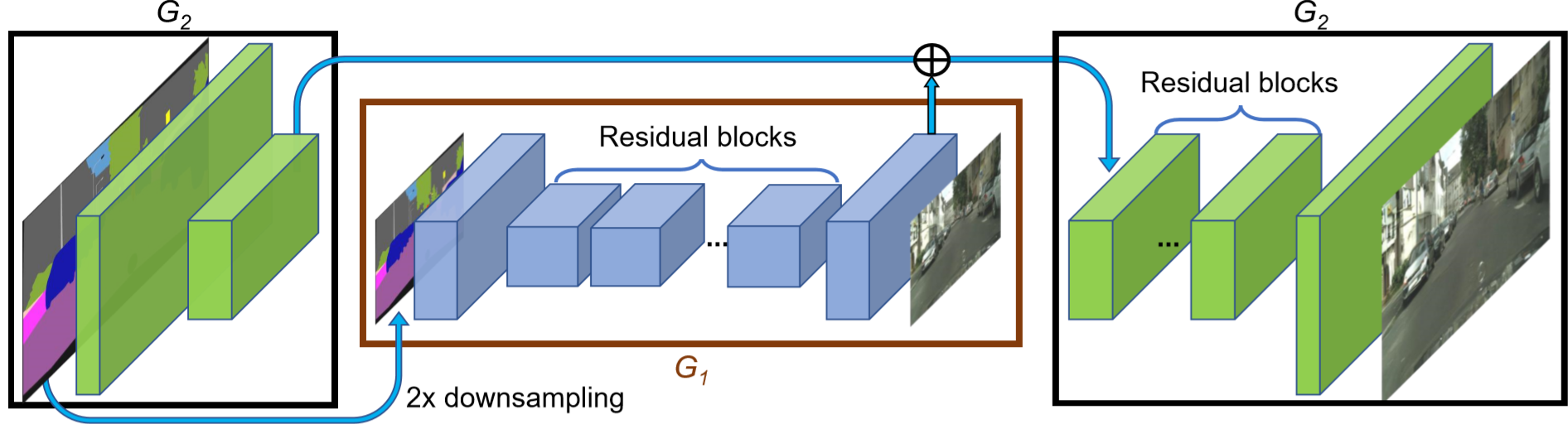}}
\caption{Architecture of the generators \cite{46}.}
\label{fig4}
\end{center}
\end{figure}

The main objective of our approach is to improve the accuracy of segmentation and generate accurate boundaries of building footprints based on deep learning and Pix2Pix networks. The proposed network benefits from the architecture of Pix2Pix network having two generators with almost the same structures, but two different discriminators to improve the accuracy of segmented objects in boundaries. As shown in Fig. \ref{fig1}, majority of the classical algorithms are not able to predict the boundaries, well. Thus, this paper tries to provide a solution by introducing G2G network. To predict the boundaries precisely, building footprint segmentation should consider the following distinct goals; 
\begin{itemize}
\item[(i)] Localization Information.
\item[(ii)] Similarity of the shape and boundaries of a segmented building to ground truth
\end{itemize} 
As mentioned before, current networks are able to segment where the buildings are, but they are not able to segment the shapes and edges of segmented buildings. Therefore, boundaries and shapes of the buildings should be taken into consideration, with special focus on how boundaries are drawn. If we assume that the most crucial information of an object are the object boundaries, adding the extracted contours of each object in the ground truth to the corresponding object in the ground truth would certainly improve the shape and boundaries of a segmented building for further research. As seen in Fig. \ref{fig2}, contours were added again to objects of the ground truths after extracting them using the Open CV library.
 \\
To go further, a Conditional Generative Adversarial Network (CGAN) called Pix2Pix network was chosen. The network is applicable for image to image translation aims including photos synthesis from label maps, objects reconstruction from edge maps, and images colorization \cite{35,46}.\\ Pix2Pix network is equipped by two parts (Fig. \ref{fig3}), Generator (G) and Discriminator (D). According to the aim, the generator converts satellite images into building maps, and the discriminator distinguishes real images from fake ones. Although, the Pix2Pix network is generalized for many different tasks without modifying the loss function, there are many issues regarding a successful training of Pix2Pix. Therefore, current researches focus on improving the training of Pix2Pix.
\\
 \\
In a recent effort, where Pix2Pix was used to synthesize an image from semantic labels \cite{46}, blurry images were produced. The study claims that the adversarial training of Pix2Pix network might be prone to failure and blurry images for high-resolution image generation task, which was attributed to unstable training. Thus, they attempted to improve Pix2Pix using two generators (Fig. \ref{fig4}) and multi-scale discriminators.
 \\
%Another prominent papers that utilize CGAN in the field of inferring contours from images uses a normal GAN as  opposed to the discriminatory structure of Pix2Pix \cite{54}. This paper claims that the Pix2Pix discriminator can help other networks to generate beautiful textures, but in this field, it leads to many break edges for a single contour of the object. 
Another study utilized CGAN in the field of inferring contours from images where normal GAN was used as opposed to the discriminatory structure of Pix2Pix \cite{54}. The study was done based on the idea that although the Pix2Pix discriminator can help other networks to generate satisfactory textures, in this case, it results in many break edges for a single contour of the object.
\subsection{Architecture of G2G network}
\begin{figure}[ht]
\begin{center}
\subfigure[Image type 1]{
\resizebox*{3.7cm}{!}{\includegraphics{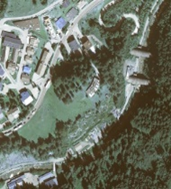}}}\hspace{3pt}
\subfigure[Image type 2 ]{
\resizebox*{3.7cm}{!}{\includegraphics{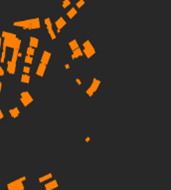}}}
\subfigure[Image type 3]{
\resizebox*{3.6cm}{!}{\includegraphics{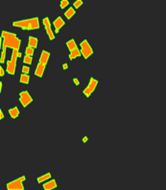}}}
\caption{Three types of images used in G2G network. }
\label{fig5}
\end{center}
\end{figure}

Considering the two major findings \cite{46,54} explained above, we concluded that the architecture of the G2G network should fulfill the following properties;
\\
\begin{itemize}
\item[1.] The structure of the Pix2Pix discriminator should be extended in order to distinguish contours as an indicator to discriminate fake and real images. The similarity of the shapes and edges of a segmented building to the ground truth can be specified with the relevant contours which play remarkably well their role as indicators.
\\
\item[2.] Two generators should be used, each having their own specified goal to focus individually on providing accurate localization and boundary features, respectively. 
\end{itemize}
In addition, three types of images (Fig. \ref{fig5}) should be defined to design the new architecture each following their own goals:\\

Image type 1: Satellite images 

Image type 2: Corresponding ground truths of satellite images 

\begin{figure}[ht]
\begin{center}
\resizebox*{12.5cm}{!}{\includegraphics{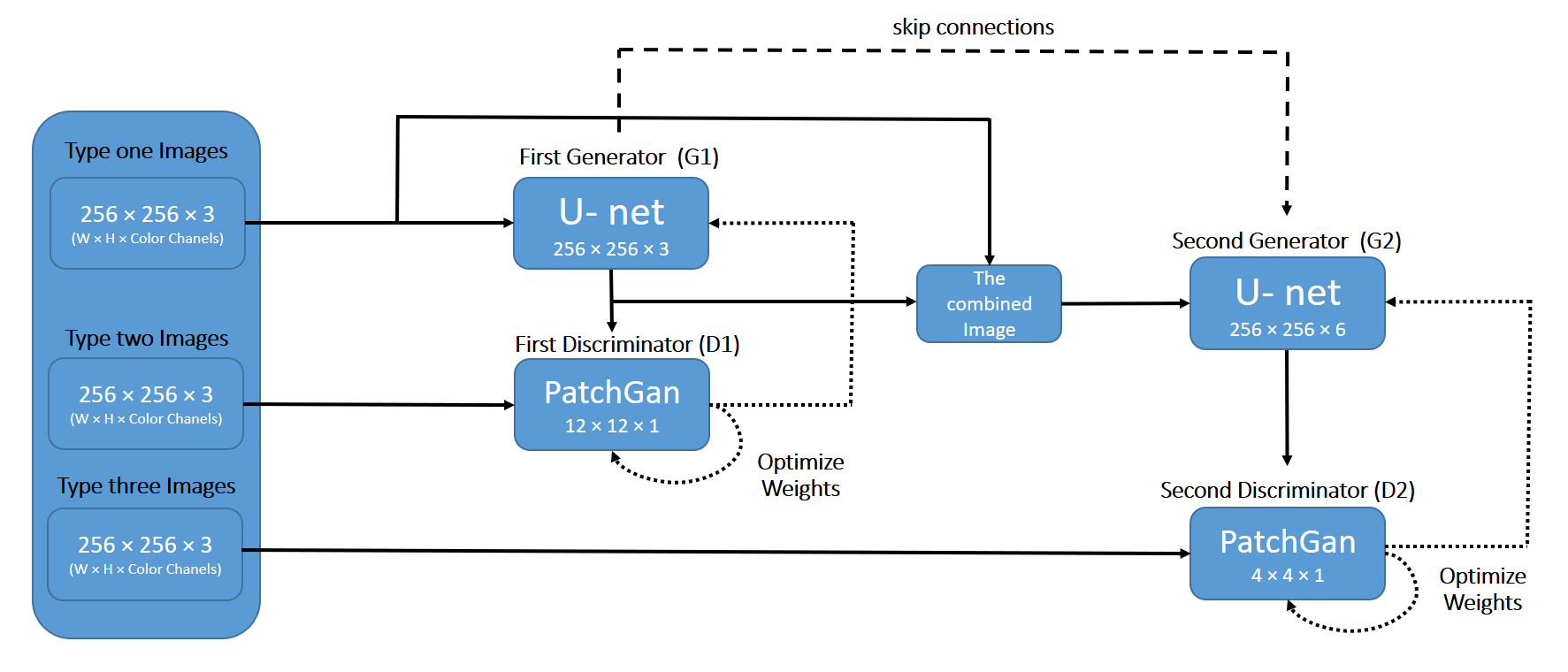}}
\caption{Architecture of G2G network.}
\label{fig6}
\end{center}
\end{figure}

Image type 3: Extracted contours overlaid on the corresponding ground truths 
\\
 \\
Accordingly to the aforementioned properties of G2G network architecture and three available types of images, the architecture of G2G network can be illustrated as follow (Fig \ref{fig6}).
\subsubsection{Architecture of the G2G generators}
\begin{figure}[ht]
\begin{center}
\resizebox*{12.5cm}{!}{\includegraphics{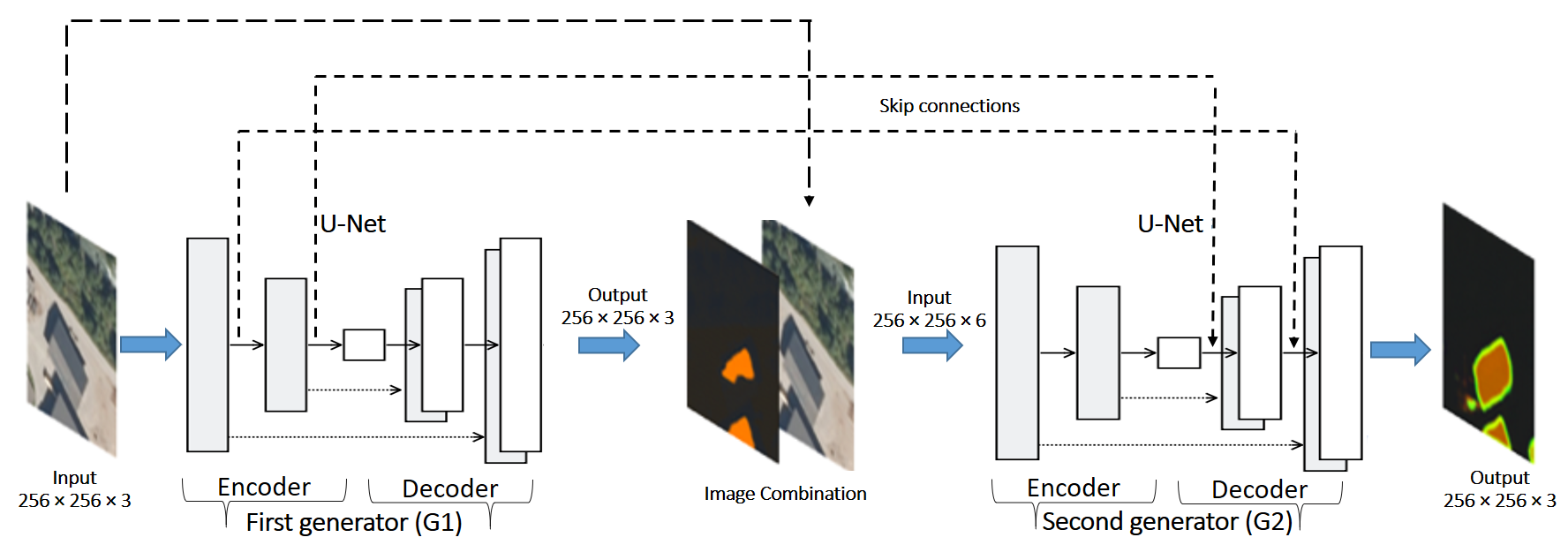}}
\caption{Architecture of the G2G generators.}
\label{fig7}
\end{center}
\end{figure}

\begin{figure}[ht]
\begin{center}
\resizebox*{12.5cm}{!}{\includegraphics{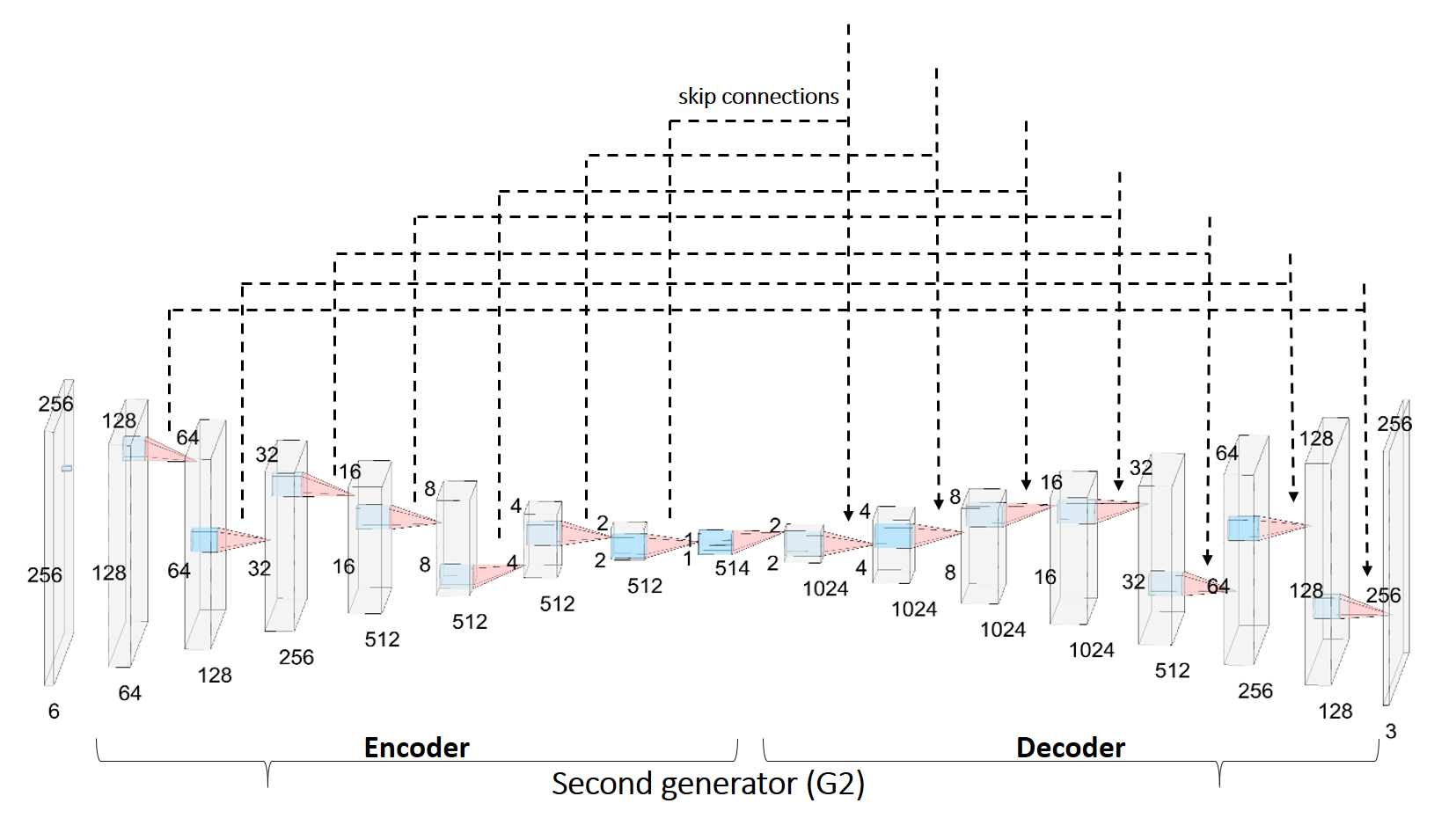}}
\caption{Structure of the second generator.}
\label{fig8}
\end{center}
\end{figure}

In this study, the global generator was decomposed into two sub-generators: G1, which uses Image 1 as an input and Image 2 as a target, with an objective to detect where the buildings are located, accomplishing the first goal. G2 that utilizes the combination of Image 1 and the first generator's output as an input, and Image 3 as a target. This sub-generator performs as a post-processing algorithm to correct the result of the first generator and at the meantime takes over the edges and contours of the objects, attaining the second goal. An individual discriminator was developed for each generator to focus more efficiently on their main allocated goal. The architecture of the G2G generators is depicted in Fig. \ref{fig7}. \\
 \\
The generators were built based on the architecture of the U-Net structure, as one of the most common and effective structures in the field of segmentation. It consists of two components: one encoder and one decoder with skip connections. In contrast to the first generator of G2G network, which has the input size of $256\times256\times3$ (height, width, channels), the second generator (Fig. \ref{fig8}) has the input size of $256\times256\times6$, representing a combination of the image generated by the first generator and the corresponding satellite image. The first generator was connected to the second generator by adding the skip connections from the initial layers of the first generator to the end layers of the second generator. \\
 \\
\subsubsection{Architecture of the G2G discriminators}
\begin{figure}[ht]
\begin{center}
\subfigure[]{
\resizebox*{10cm}{!}{\includegraphics{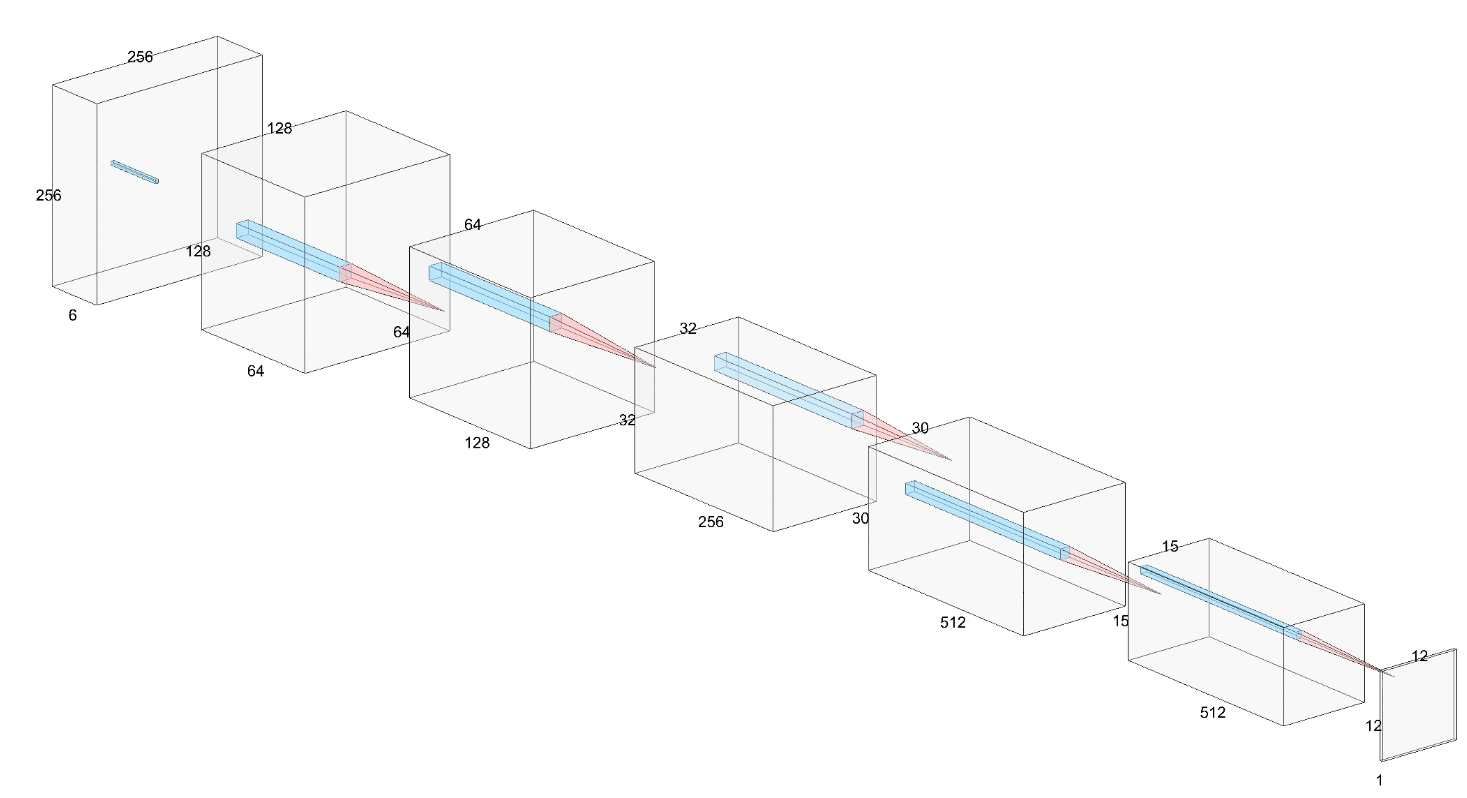}}}\hspace{3pt}
\subfigure[]{
\resizebox*{10cm}{!}{\includegraphics{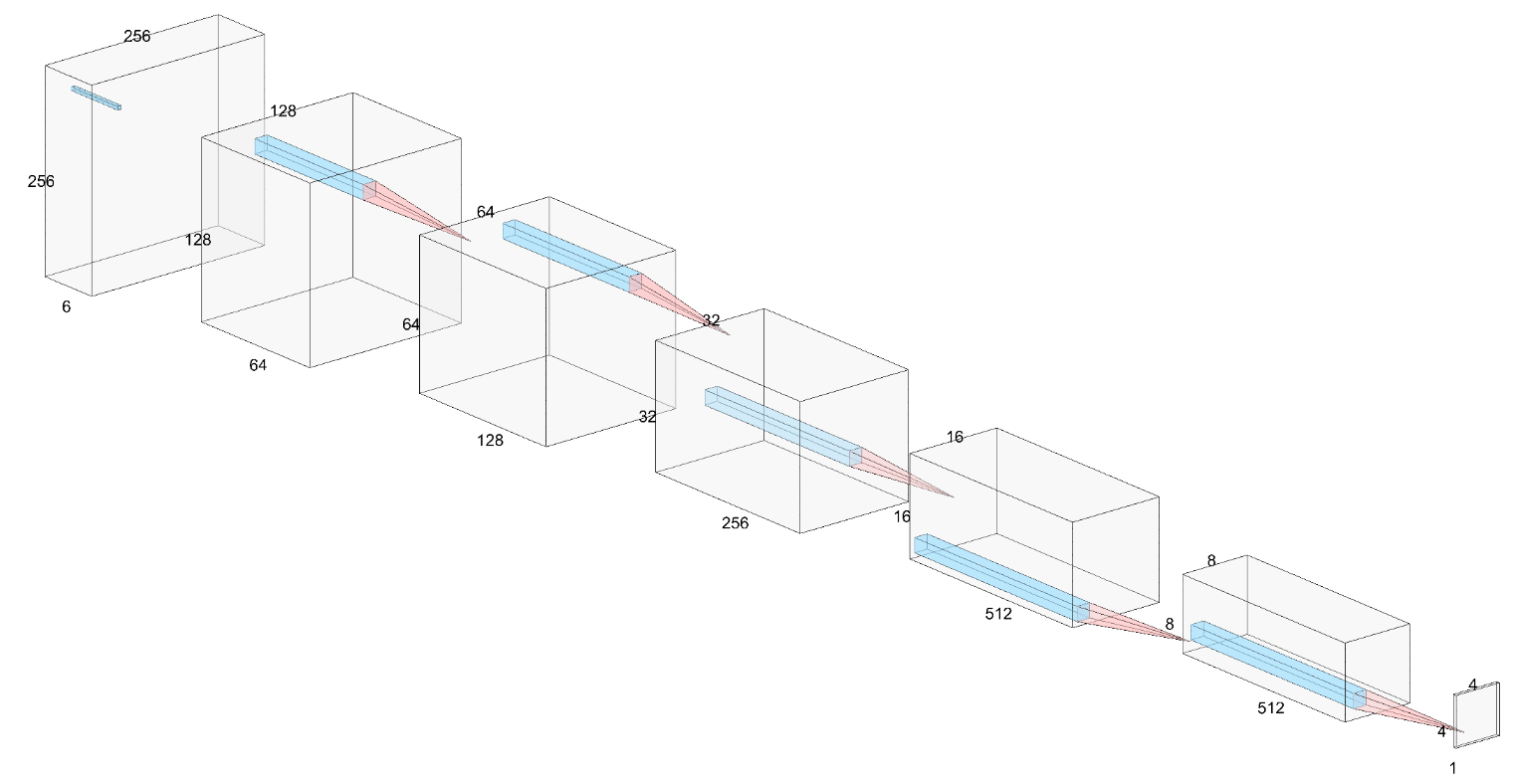}}}
\caption{Structures of the two discriminators, (a) Structure of the first discriminator (D1), (b) Structure of the second discriminator (D2).}
\label{fig9}
\end{center}
\end{figure}

Two different discriminators with two distinct network structures were designed to operate on identical image scales which are denoted as D1 and D2 for the first and second discriminators, respectively. Due to using Zero Padding layer in the structure of the original Pix2Pix discriminator, a lot of information is lost and the structure of the two proposed discriminators is built without any Zero Padding layer Fig. \ref{fig9}(a). The first layer of D1 is a concatenated layer of inputs followed by the three convolutional layers consisting of convolutional layer, batch normalization, and leaky rectified linear unit (leaky ReLU) as an activation function which has a small slope for negative values, instead of zero values. \\
The output of the third convolutional layer passes through the fourth layer, which is a series of convolutional and batch normalized layers without leaking ReLU. Then, the output of this layer was fed to the next one, including max pooling and convolutional layers. The output size of the final layer is $12\times12\times1$, which corresponds to the $28\times28$ patch of the input image, in contrast to the original Pix2Pix discriminator where each pixel of the output discriminator ($30\times30$) corresponds to the $70\times70$ patch of the input image. As shown in Fig. \ref{fig9}(b), in the second discriminator structure, the first six layers have the same structure as the first four layers of D1 but the output of the fifth layer is reduced by half and followed by that, it is fed to the sixth layer. To decrease variance and complexity of computation, a max pooling layer was added at the end of the D2 structure in G2G network, to extract low-level and important features including edges from the neighborhood. \\
 \\
The output size of the final layer is  $4\times4\times1$, which corresponds to the $9\times9$ patch of the input image. Unlike the first discriminator, where each pixel of the output ($12 \times12$ image) is relevant to the $28\times28$ patch of the input image, in the second discriminator, each pixel of the output corresponds to the $9\times9$ patch of the input image resulting in more accurate object segmentation.
\subsection{Training phase of G2G network}
\begin{figure}[ht]
\begin{center}
\resizebox*{12.5cm}{!}{\includegraphics{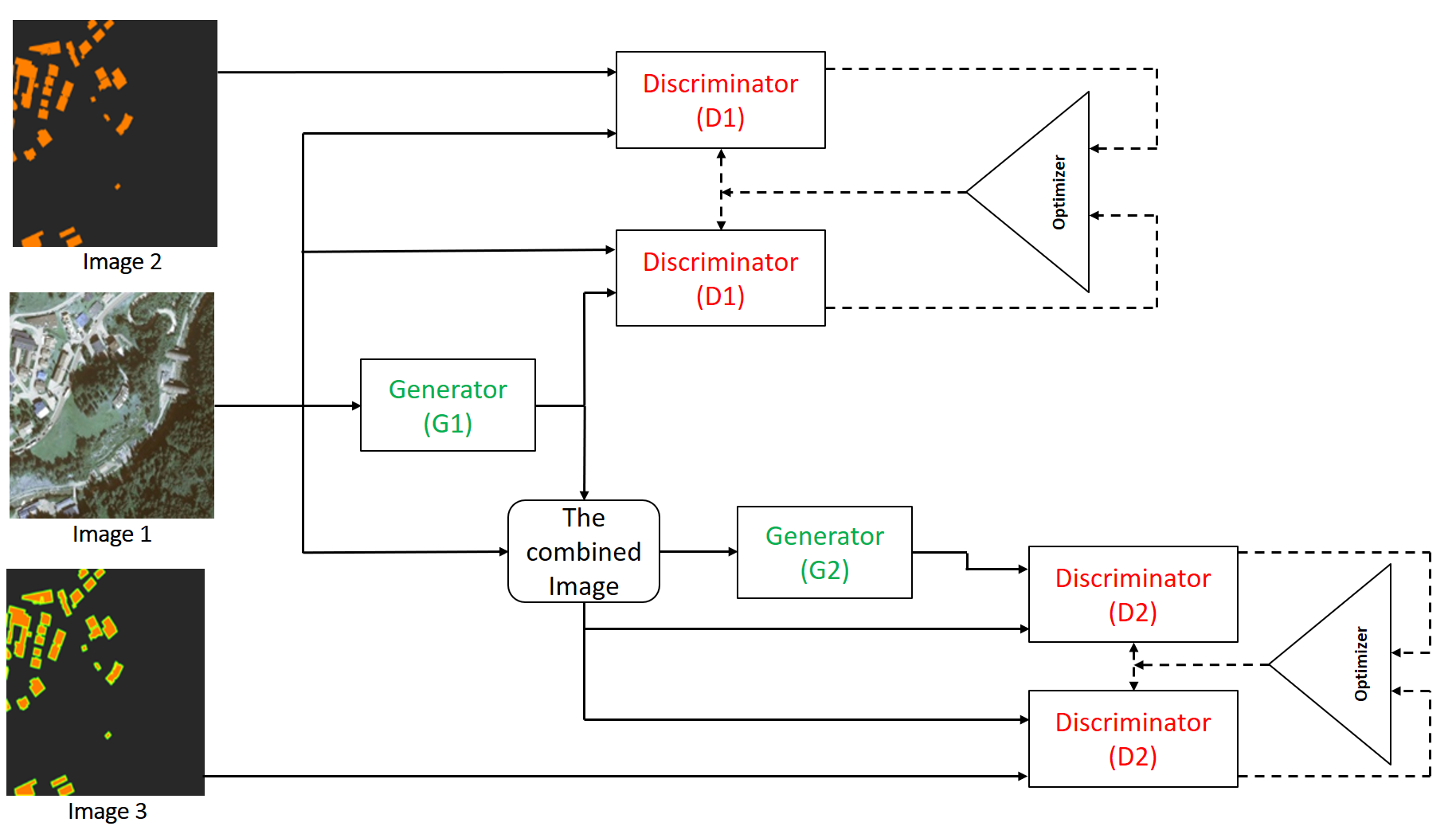}}
\caption{Flowchart of adjusting the weights of G2G.}
\label{fig10}
\end{center}
\end{figure}

To adjust the weights of G2G, two steps are performed (Fig. \ref{fig10}). In the first step, the first discriminator takes the input (Image 1)/ target (Image 2) and then input (Image 1)/ output (the generated output of the first generator) pairs and makes its estimate on how realistic they look. Next, based on the differences, the weights of the first discriminator will be adjusted. At the end, the combination of the first generator's output with the corresponding satellite image (Image 1) passes through the second generator as input.  Thereafter, the output of the second generator will be calculated and inserted into the second discriminator, which compares the input (the combined image)/ target (Image 3) then, the input (the combined image)/output (the generated output of the second generator) pairs and makes its estimate on how realistic they look in order to adjust the weights of the second discriminator. \\
 \\
In the second step, the weights of the generators were adjusted according to the outputs of the discriminator and differences between the objectives and the outputs. In this study, the same objective function of Pix2Pix network (1), was employed because of its efficiency for segmentation tasks;
\[G^*=\arg\min_G\min_D\Big(\mathcal{L}_{cGAN}(G,D)+\lambda\mathcal{L}_{L1}(G)\Big),\] where the loss function $L1$ called Manhattan distance \cite{Man} measures the standard distance between the generated output and the target.
\subsection{Training strategy}
Following the definition of the training phase for training G2G network, a strategy was developed to train the network, which includs the following two steps:
\\
\begin{itemize}
\item[1.]	Simultaneous training of the generators and discriminators with a nearly high learning rate of $10\textrm{e}-3$ recommended in previous studies on Pix2Pix network \cite{35} and 200 epochs.
\\
\item[2.]	Training the second generator and discriminator with a nearly low learning rate of $10\textrm{e}-6$ and 200 epochs.
\end{itemize}
\section{Experimental results}
This section is devoted to discuss the dataset used in the training phase of G2G network and the evaluation criteria which will be selected to score our approach and results.
\subsection{Dataset}
\begin{figure}[ht] 
\begin{center}
\subfigure[Step 2]{
\resizebox*{3.7cm}{!}{\includegraphics{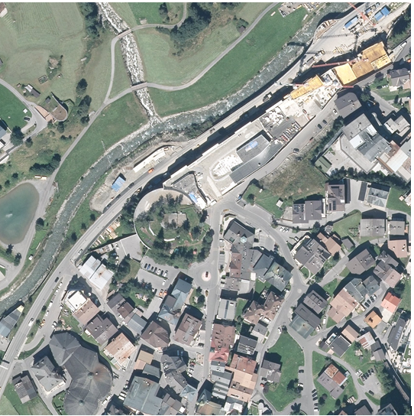}}}\hspace{3pt}
\subfigure[Step 3 ]{
\resizebox*{3.2cm}{!}{\includegraphics{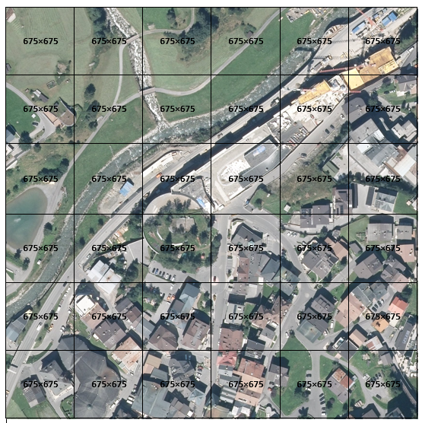}}}
\subfigure[Step 4]{
\resizebox*{2.7cm}{!}{\includegraphics{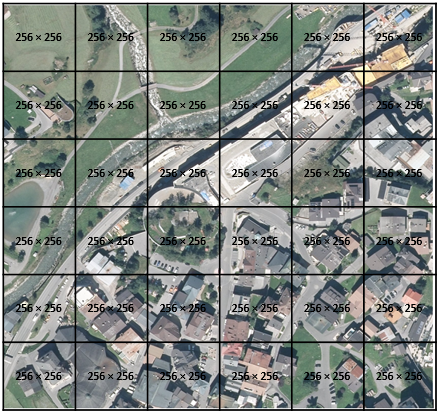}}}
\caption{Three last steps of the planned strategy for creating training, validation and test datasets.}
\label{fig11}
\end{center}
\end{figure}

The Orthofoto Tirol dataset \cite{http} was used to investigate the performance of the present approach. The dataset consists of two categories; satellite images and ground truths. Main characteristics in terms of Number, size, and type of the images are as presented in Table 1. 
\begin{table}[ht]\label{tab1}
\begin{center}
\begin{small}
\caption{Characteristics of the Orthofoto Tirol dataset.}\vspace*{0.1in}
\begin{tabular}{ccccccccccc}
\hline
\noalign{\smallskip}
&& Name of the category && Number of images && Size of images && Type of images\\
\noalign{\smallskip}\hline\noalign{\smallskip}
$1$&& Satellite Images&& $ 21,076 $&&$4053\times4053$&& PNG \\
$2$&& Ground truths && $ 21,076 $&& $4053\times4053$&& PNG\\
\hline
\end{tabular}
\end{small}
\end{center}
\end{table} \\
Since segmentation models are using every single pixel of images with specific input sizes, all pixels of an image are of equal importance. In case that the number of the pixels have to be reduced due to the limited size of a model input, the image will be much smaller and this results in loss of pixels. Therefore, it is important to retain more information of images by resizing them without losing the pixels. 
Images extracted from the Orthofoto Tirol dataset needed to be cut into $256\times256$, since the G2G is built based on Pix2Pix network, which only takes $256\times256$ images. In the first attempt, the size of the satellite and corresponding ground truth images was scaled down all at once to $256\times256$, which led to loss of 16.361.273 pixels. Since each pixel matters in precise detection of the edges of buildings, to avoid loosing of the main pixels, a strategy with the following steps was planned: \\
\begin{itemize}
\item[1.]	Images with high building density were selected from the entire dataset.
\\
\item[2.]	The size of each selected image ($4053\times4053$ pixels) was changed to $4050\times4050$ pixels, Fig. \ref{fig11}(a).
\\
\item[3.]	All images from the previous step were cropped to the size of $675\times675$ pixels, Fig. \ref{fig11}(b).\\
\item[4.]	All cropped images were scaled down to $256\times256$ pixels, hence not a big deal of information was lost. Number of 390.089 pixels were lost, which is far fewer than 16.361.273 pixels, Fig. \ref{fig11}(c). \\
\end{itemize} 
This strategy was performed to create datasets for training, validation, and testing purposes, where the testing dataset is used to assess the accuracy of the model and the validation dataset is needed to optimize the model during the training period, in order to minimize overfitting and fine-tuning of the hyper-parameters of the model. Properties of each created dataset are indicated in Table 2.
\begin{table}[ht]\label{tab2}
\begin{center}
\begin{small}
\caption{Characteristics of the training, validation and test datasets derived from the Orthofoto Tirol dataset.}\vspace*{0.1in}
\begin{tabular}{cccccc}
\hline
\noalign{\smallskip}
Dataset&& Selected images& Cropped images for each image & Entire images\\
\noalign{\smallskip}\hline\noalign{\smallskip}
Training dataset&& $191$&$ 36 $&$6876$ \\
Validation dataset&& $21$ & $ 36 $& $756$\\
Test dataset&& $22$& $36 $& $792$\\
\hline
\end{tabular}
\end{small}
\end{center}
\end{table} \\
\subsection{Evaluation criteria}
\begin{figure}[ht]
\begin{center}
\resizebox*{4.5cm}{!}{\includegraphics{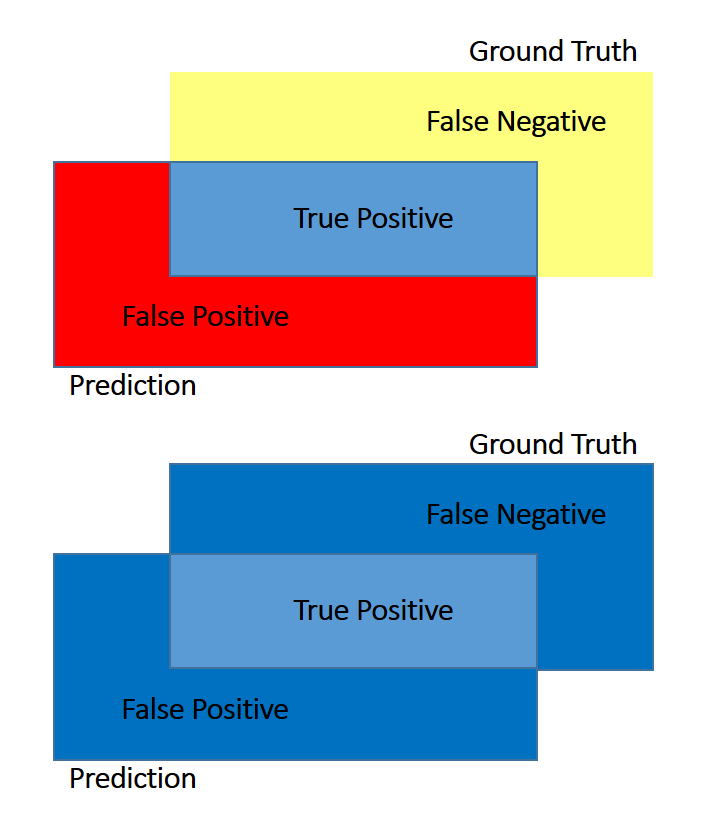}}
\caption{An Intersection over Union example for the Dice similarity coefficient. }
\label{fig54}
\end{center}
\end{figure}

Evaluation of the quality of a segmentation model is essential specifically for image processing in security cases such as autonomous vehicles. Although, there are many evaluation criteria developed for evaluating of segmentation models, there is no conclusive effective technique for selecting the best criteria \cite{55,56}.
Dice similarity coefficient is one of the common criteria  which determines the similarity between the results of segmentation and its corresponding ground truths in a slightly different way \cite{77}. Considering the ground truth and prediction boxes (Fig. \ref{fig54}), the overlapping area between the two mentioned boxes showed in blue rectangle display where the pixels of the ground truth and prediction match, and it is called true positive (TP). In addition, the red region includes the pixels called false positives (FP).\\
 \\
Yellow region in Fig. \ref{fig54}, possesses the pixels, which are missed from segmentation and called false negatives (FN). When the overlapping area completely covers the union area, the segmentation is flawless, the value of the Intersection over Union (IoU) criteria is 1, and the values of FP, TP, and FN are zero. Therefore, the IoU criteria formula was rewritten as follow \cite{78}:
\begin{equation}
IoU=\dfrac{TP}{(TP+FP+FN)}
\end{equation} \\
Accordingly, Dice similarity coefficient was calculated as follow \cite{78}:\\
\begin{equation}
Dice=\dfrac{2\times TP}{(TP+FP)+(TP+FN)}
\end{equation} 
%\end{itemize}
In what follows, we receive hand from the following four evaluation metrics inspired by the outstanding work \cite{32} to evaluate the efficiency of the G2G segmentation network.
The metrics are extracted based on pixel accuracy and IoU metric, assuming:\\
	$n_{cl}$: the number of classes,\\
	$t_i$: the total number of pixels in class i,\\
	$n_{ij}$: the number of pixels of class we predicted to belong to class j, \\
	$n_{ii}$: the number of correctly classified pixels (true positives),\\
	$n_{ij}$: the number of pixels wrongly classified (false positives),\\
	$n_{ji}$: the number of pixels wrongly not classified (false negatives).\\
	\begin{itemize}
\item[I) ] Pixel Accuracy (PA): The overall accuracy that is calculated as follow  \cite{32}: 
\begin{equation}
PA=\dfrac{\sum\limits_i n_{ii}}{\sum\limits_i t_{i}}.
\end{equation}
\item[II)] Mean Accuracy (MA): The average accuracy among all the classes related to ground truths. This metric is calculated as follow\cite{32}: 
\begin{equation}
MA=\dfrac{1}{n_{cl}}\sum\limits_i \dfrac{n_{ii}}{t_{i}}.
\end{equation}
\item[III)]	Mean Intersection over Union (MIoU): Commonly used for semantic segmentation performance evaluation which is calculated based on Dice similarity as follow \cite{32}:
\begin{equation}
MIoU=\dfrac{1}{n_{cl}}\sum\limits_i \dfrac{n_{ii}}{t_{i}+\sum\limits_{j}n_{ji}-n_{ii}}.
\end{equation}
\item[IV)]	Frequency weighted intersection over union (FWIoU): The metric considers the number of data points in each class  \cite{32} which is calculated as below:
\begin{equation}
FWIoU=\Big(\sum\limits_k t_k\Big)^{-1}\sum\limits_i \dfrac{t_i n_{ii}}{t_{i}+\sum\limits_{j}n_{ji}-n_{ii}}.
\end{equation}\end{itemize}
Unlike the MIoU and FWIoU, the other two aforementioned metrics are not susceptible to unbalanced datasets \cite{55,56}.
\subsection{Evaluation results}
\begin{figure}[ht]
\begin{center}
\resizebox*{12.5cm}{!}{\includegraphics{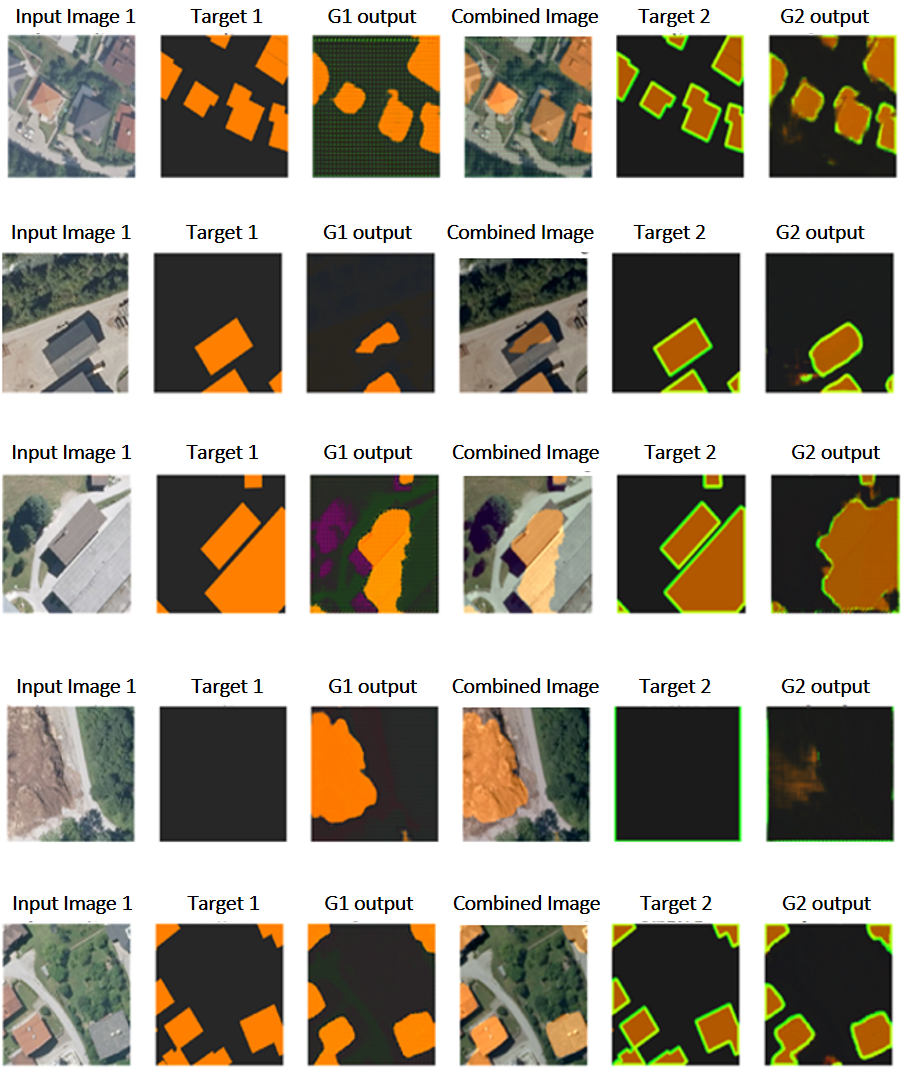}}
\caption{Results of the G2G on the validation dataset.}
\label{fig52}
\end{center}
\end{figure}

There are several strategies to evaluate the outputs of G2G. However, current study developed two different strategies to quantify the quality of the G2G's results. In the first strategy, all four evaluation metrics were used which produced more accurate results in the field of segmentation. Regarding this strategy, ground truths and images predicted by the G2G were fully compared using the four evaluation metrics.\\ Finally, to provide a global mean value of all test images, values of each metric were averaged. The second evaluation strategy was training of original Pix2Pix network with its corresponding recommended hyper-parameters which were used to train the G2G.  The Pix2Pix network was evaluated on the same test dataset using the four evaluation metrics in order to obtain a global mean value of all the test images. Following determining of the global mean values of the G2G and Pix2Pix network, they were compared to see which network outperforms the other.\\
Results driven from the two strategies (Table 3) demonstrate that the G2G performed more efficient than Pix2Pix in regards to representing a large margin for all the evaluation metrics. Furthermore, comparison between the mean IoU of G2G and the recent multi-task learning study \cite{11} (Table 3) verifies the superiority of the present network. \\
\begin{table}[ht]\label{tab3}
\begin{center}
\begin{small}
\caption{Comparison between G2G, Pix2Pix and multi-task learning \cite{11} networks in terms of the percentage values of the four evaluation metrics.}\vspace*{0.1in}
\begin{tabular}{cccccccc}
\hline
\noalign{\smallskip\noindent}
Metrics&& G2G  && Pix2Pix &&Multi-task \cite{11}\\
\noalign{\smallskip}\hline\noalign{\smallskip}
Mean pixel accuracy&& $0.96$&&$ 0.89$ && $0.95$\\
Mean accuracy&& $0.89$ && $ 0.74$&&$-$\\
Mean IoU&& $0.83$&&$0.65$&&$0.70$\\
Mean frequency weighted IU&& $0.90$&& $0.82$&&$-$\\
\hline
\end{tabular}
\end{small}
\end{center}
\end{table} \\
In line with the applied strategies, G2G network was tested on validation dataset and the second generator of G2G network acted as a post-processing algorithm (Fig. \ref{fig52}) trying to improve the output of the first generator and making predictions that looked like ground truth from the perspective of boundaries and edges.\\
\section{Conclusion}
Current study developed a new network called G2G, based on Pix2Pix network, to solve the problem of imprecise boundaries of the building footprint segmentation, which was mainly conducted through conversion of high-resolution satellite images into maps of the buildings. This novel network is based on employing two generators with certain individual purposes to improve the accuracy of building footprint segmentation. The network preserves the local and boundary information which gives it the opportunity to achieve a high accuracy in the segmentation analysis. Different criteria verify the superiority of G2G network against other pre-existing networks.
%\begin{acknowledgements}
%\end{acknowledgements}



\bibliographystyle{spmpsci}
\bibliography{bibli31}
%\begin{thebibliography}{}
%%
%% and use \bibitem to create references. Consult the Instructions
%% for authors for reference list style.
%\end{thebibliography}

\end{document}